\documentclass{article}

\usepackage{microtype}
\usepackage{graphicx}
\usepackage[off]{svg-extract}
\usepackage{multirow}
\usepackage{subcaption}

\usepackage{booktabs} 

\usepackage{hyperref}

\usepackage[accepted]{icml2025}

\usepackage{amsmath}
\usepackage{amssymb}
\usepackage{mathtools}
\usepackage{amsthm}

\usepackage[capitalize,noabbrev]{cleveref}

\theoremstyle{plain}

\theoremstyle{definition}

\theoremstyle{remark}

\usepackage[textsize=tiny]{todonotes}

\icmltitlerunning{Continual Generalized Category Discovery Using Extreme Value Theory and Proxy Anchors}

\begin{document}

\twocolumn[
\icmltitle{Proxy-Anchor and EVT-Driven Continual Learning Method for Generalized Category Discovery}




\icmlsetsymbol{equal}{*}

\begin{icmlauthorlist}
\icmlauthor{Alireza Fathalizadeh}{sch}
\icmlauthor{Roozbeh Razavi-Far}{sch}
\end{icmlauthorlist}

\icmlaffiliation{sch}{University of New Brunswick, Fredericton, Canada}

\icmlcorrespondingauthor{Alireza Fathalizadeh}{alireza.fathalizadeh@unb.ca}

\icmlkeywords{Machine Learning, ICML}

\vskip 0.3in
]



\printAffiliationsAndNotice{}  
\newlength{\defaultabovedisplayskip}
\newlength{\defaultbelowdisplayskip}
\setlength{\defaultabovedisplayskip}{\abovedisplayskip}
\setlength{\defaultbelowdisplayskip}{\belowdisplayskip}
\begin{abstract}
Continual generalized category discovery has been introduced and studied in the literature as a method that aims to continuously discover and learn novel categories in incoming data batches while avoiding catastrophic forgetting of previously learned categories. A key component in addressing this challenge is the model's ability to separate novel samples, where Extreme Value Theory (EVT) has been effectively employed. In this work, we propose a novel method that integrates EVT with proxy anchors to define boundaries around proxies using a probability of inclusion function, enabling the rejection of unknown samples. Additionally, we introduce a novel EVT-based loss function to enhance the learned representation, achieving superior performance compared to other deep-metric learning methods in similar settings. Using the derived probability functions, novel samples are effectively separated from previously known categories. However, category discovery within these novel samples can sometimes overestimate the number of new categories. To mitigate this issue, we propose a novel EVT-based approach to reduce the model size and discard redundant proxies. We also incorporate experience replay and knowledge distillation mechanisms during the continual learning stage to prevent catastrophic forgetting. Experimental results demonstrate that our proposed approach outperforms state-of-the-art methods in continual generalized category discovery scenarios. Our code is publicly
available at 
\href{https://github.com/NumOne01/CATEGORIZER}{https://github.com/NumOne01/CATEGORIZER}.

\end{abstract}
\begin{figure*}
    \centering
    \includesvg[width=\linewidth,inkscapelatex=false]{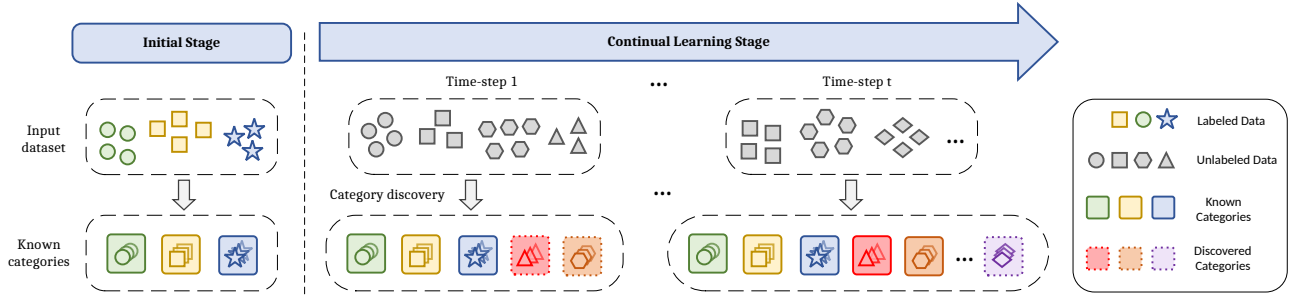}
    \caption{The general presentation of the Continual Generalized Category Discovery (CGCD) setting. In the initial stage, a labeled dataset is provided to train the initial model. After the initial stage, the model enters the continual learning stage, in which no labeled data is provided. The input data in this stage can contain samples belonging to novel or previously known categories. The model is expected to discover potential novel categories in this unlabeled data and integrate them into the model without compromising the performance of previous categories and making assumptions about the number of novel categories.}
    \label{fig:scenario}
\end{figure*}

\section{Introduction}
Most traditional machine learning algorithms operate under the closed-world assumption, in which the training and test data are drawn from the same label and feature spaces and no new class is introduced to the model during the test phase. However, in a more practical scenario, the training data lacks complete knowledge of the world and unknown classes may emerge during the test phase. A crucial problem is that a model that operates under the closed set assumption usually makes high-confidence predictions for these novel samples. This is particularly problematic in critical systems like autonomous driving, where misclassification can cause serious harm, requiring the model to detect novel samples and integrate potential novel classes into the knowledge base of the model.

Humans have the ability to identify new entities and group them without prior knowledge, while also recognizing them upon subsequent encounters., e.g., one can see new types of bird species that they did not see before, but they can still group and categorize them once they see them and add this new information to their knowledge base without affecting what they know about other types of birds, i.e., no forgetting. This has inspired a scenario, called \underline{C}ontinual \underline{G}eneralized \underline{C}ategory \underline{D}iscovery (CGCD) \cite{zhang2022grow, wu2023metagcd, kim2023proxy, zhao2023incremental}, in which a model is trained on an initially labeled dataset and after this initial stage, the model is only introduced to unlabeled data and is expected to detect and discover potential novel categories in the data and integrate them into the model without compromising the performance of previously learned tasks.

The problem of continual generalized category discovery can be decomposed into three subtasks:
1) \textbf{Novelty detection}: Detecting samples that do not belong to any previously learned and known categories \cite{geng2020recent, liu2020energy, yang2024generalized}, 2) \textbf{Category discovery}: Identifying potential novel categories in an unlabeled dataset \cite{han2021autonovel, fini2021unified, vaze2022generalized}, and 3) \textbf{Continual learning}: Integrating newly discovered categories into the model without catastrophic forgetting of previously learned categories \cite{rebuffi2017icarl, rolnick2019experience, wang2024comprehensive}.
Almost any method that solves each of these sub-tasks can be combined to handle the CGCD scenario, however, this might result in sub-optimal solutions as each method is trying to solve a different task, and balancing these competing objectives effectively
is challenging \cite{zhang2022grow}. A unified framework tailored to this problem is then necessary to achieve a promising balance. 

A crucial component of such a framework is the ability to clearly distinguish between known and unknown samples. To achieve this, we leverage Extreme Value Theory (EVT), which has demonstrated its effectiveness in addressing the open-set recognition problem \cite{bendale2016towards, rudd2017extreme, geng2020recent}.  EVT is a statistical framework for modeling extreme deviations in data by analyzing the tail ends of the distribution \cite{coles2001introduction}. In this context, EVT is used to model the margin distribution of each proxy, which is a representative of a class and learned as a part of the network parameters \cite{kim2020proxy}, relative to other samples and define boundaries around proxies to reject unknown samples by developing a probability of inclusion function. Building on this, we propose a novel loss, called $evt$ loss, which is derived from the EVT analysis of each proxy. In addition to preparing the model for unknown rejection, this loss improves the learned representation.\newline In the context of novel class discovery, existing methods often rely on clustering techniques, which tend to overestimate the number of classes. We mitigate this issue by utilizing EVT to reduce the number of estimated novel categories by discarding redundant ones, leading to improved discovery of new classes, while minimizing the forgetting of previously learned ones. To avoid catastrophic forgetting of previously learned data, we employ commonly used methods of knowledge distillation \cite{li2017learning, wu2018memory, hou2019learning} and experience replay \cite{chaudhry2019tiny, rolnick2019experience, lin2023pcr, zheng2024selective}.
In summary, our contribution can be summarized as follows:
\begin{itemize}
    \item We propose a novel approach, called proxy-an\underline{c}hor \underline{a}nd EV\underline{T}-driven continual-l\underline{e}arnin\underline{g} meth\underline{o}d fo\underline{r} general\underline{iz}ed cat\underline{e}gory discove\underline{r}y (CATEGORIZER). Extensive evaluations on multiple datasets demonstrate that our proposed method outperforms state-of-the-art approaches in the CGCD setting.
    \item We introduce a novel loss function called $evt$ loss, which is derived from proxy anchors \cite{kim2020proxy} and extreme value theory. This loss outperforms deep metric learning methods used in similar methods.
    \item  We propose a novel approach to mitigate over-estimating the number of novel categories in the discovery phase. This has been done by means of extreme value theory, which boosts the performance of the model in discovering categories as well as minimizing the forgetting of previously learned categories.
\end{itemize}

\section{Related Works}
\textbf{Novelty detection} aims to identify samples from novel classes not encountered during the training phase. 
Methods for novelty detection can be categorized into two main groups of open-set recognition (OSR) \cite{geng2020recent, chen2021adversarial, zhou2021learning} and out-of-distribution (OOD) detection \cite{liu2020energy, sun2021react, kim2024neural}. Despite having small differences, these methods share the goal of detecting samples from unknown categories.

\textbf{Category discovery} methods try to identify novel categories within unlabeled data \cite{han2021autonovel, zhao2021novel, zhong2021neighborhood}, where the unlabeled data only contains novel samples. Recent advancements have addressed more realistic scenarios through Generalized Category Discovery (GCD), where unlabeled data includes both previously known and novel classes \cite{fini2021unified, vaze2022generalized, zhao2023learning}.

\textbf{Continual learning} methods aim to address the challenge of catastrophic forgetting, enabling models to retain previously learned knowledge, while adapting to new tasks. Existing approaches can be broadly categorized into replay methods \cite{chaudhry2019tiny, rolnick2019experience, lin2023pcr, zheng2024selective}, regularization methods \cite{ zenke2017continual, lopez2017gradient, wu2018memory, wistuba2023continual}, meta-learning methods \cite{finn2017model, gupta2018meta, ostapenko2019learning}, architecture-based methods \cite{rusu2016progressive, yoon2017lifelong, hu2023dense}, and hybrid methods \cite{li2019learn, douillard2020podnet, wu2021incremental}.

\textbf{Continual generalized category discovery} methods aim to identify novel categories in unlabeled data, which may include both novel samples and samples from previously known classes. These methods tackle the challenge of discovering new categories incrementally, while retaining knowledge of existing ones.  
Grow and Merge (GM) \cite{zhang2022grow} introduces a framework with two models: a static model that preserves knowledge of old classes and a dynamic model trained in a self-supervised manner to discover novel classes. Novel samples are identified based on their distance to the prototypes of known classes, and the two models are merged, when a new task arrives. Incremental Generalized Category Discovery (IGCD) \cite{zhao2023incremental} employs a non-parametric classifier combined with a density-based exemplar selection method that is employed to select exemplars samples as well discovering novel classes.  
Proxy Anchor (PA) \cite{kim2023proxy} uses proxies learned through proxy anchor loss \cite{kim2020proxy} to detect novel samples. A non-parametric clustering algorithm clusters and identifies new categories, while catastrophic forgetting is mitigated through experience replay and knowledge distillation. MetaGCD \cite{wu2023metagcd} adopts a meta-learning framework that leverages offline labeled data, to simulate the incremental learning process and utilizes the classic k-means algorithm for novel class discovery.

\section{Problem Definition}
This section presents the problem setting.
In the framework of Continual Generalized Category Discovery (CGCD), the model begins with an initial labeled dataset 
\(D^0_l = \{(x, y) \in \mathcal{X}_l^0 \times \mathcal{Y}_l^0\}\), where \(\mathcal{X}_l^0\) represents the labeled 
input data and \(\mathcal{Y}_l^0\) denotes the corresponding labels. This dataset is used to train a feature extractor 
\(\mathcal{F}^0 : \mathcal{X} \to \mathcal{Z}^0\), which generates embedding vectors \(\mathcal{Z}^0\), and a classifier 
\(\mathcal{C}^0 : \mathcal{Z}^0 \to \mathcal{Y}^0\), which maps the embeddings to specific classes.

Following this initial phase, no additional labeled data is provided. Instead, the model is sequentially exposed to a 
series of unlabeled datasets, \(\{ D^t_u \}_{t=1}^{T}\), where \(D^t_u = \{x \in \mathcal{X}^t_u\}\). 
Here, \(\mathcal{X}^t_u\) contains the unlabeled input data and \(T\) denotes the number of time steps. These datasets include examples from previously encountered categories as well as samples from potentially 
new categories, i.e., \(\mathcal{Y}_l \cap \mathcal{Y}_u \neq \emptyset\). No assumption is made about the presence or the number of novel categories in the unlabeled data.

The model’s task is to identify and learn new categories, while updating the feature extractor 
\(\mathcal{F}^t : \mathcal{X} \to \mathcal{Z}^t\) and classifier \(\mathcal{C}^t : \mathcal{Z}^t \to \mathcal{Y}^t\) at each 
time step \(t\). Over time, \(\mathcal{Y}^t\) encompasses all categories seen so far, i.e., \(\mathcal{Y}^t = \mathcal{Y}^{t-1} \cup \mathcal{Y}^t_{new} \), where \(\mathcal{Y}^t_{new}\) is the newly discovered categories in the current time step \(t\). Figure \ref{fig:scenario} illustrates this setting. \newline
The key challenge lies in balancing 
the catastrophic forgetting of previously learned data (stability) with the effective discovering and learning of new categories (plasticity).

\begin{figure*}[!ht]
    \centering
    \includesvg[width=\linewidth,inkscapelatex=false]{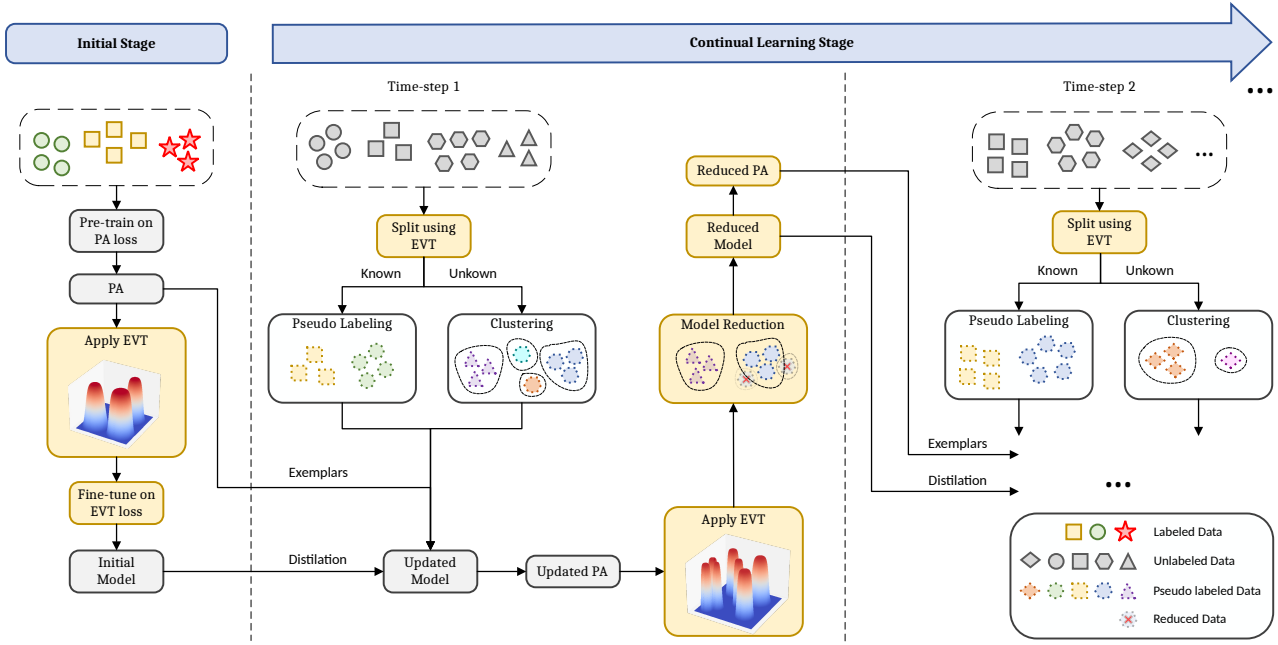}
    \caption{Overview of  CATEGORIZER. In the initial stage, the model is first pre-trained on PA loss to derive proxy anchors for different classes. Following this, the EVT analysis is applied to each proxy to compute the Weibull distribution around each proxy and devising a probability of inclusion (PSI) function that is capable of rejecting unknown samples. With the computed distributions, we fine-tune the model on our novel $evt$ loss to get the initial model. In the continual learning stage, the input data containing novel and known samples are separated by thresholding PSIs functions computed in the initial stage into known and unknown samples. Known samples are pseudo-labeled using the current model from the previous step and unknown samples are clustered. The model is updated using pseudo-labeled and clustered data, exemplars of previous categories, and distillation loss derived from the previous step model. EVT is applied to the updated model to get updated distribution, where the model is reduced and redundant proxies are discarded. This process repeats for the next steps. The yellow boxes indicate our novel contribution in the proposed scheme.}
    \label{fig:overview}
\end{figure*}
\section{Proposed Method}
Our proposed novel method is composed of two main stages: 1) the initial stage and 2) the continual learning stage. 
In the initial stage, the feature extractor \(\mathcal{F}^0\) and classifier \(\mathcal{C}^0\) are trained using the labeled dataset. 
In the subsequent continual learning stage, the process begins with a novelty detection module that divides the data into 
known and unknown sets. The unknown samples are then subjected to a novel class discovery procedure, which identifies 
coherent and distinct samples as potential new classes. Finally, the feature extractor and classifier are updated to incorporate 
the newly discovered samples. Figure \ref{fig:overview} illustrates the overview of CATEGORIZER. The following sections provide detailed explanations of the main steps involved.

\subsection{Initial Stage}
To train the feature extractor \(\mathcal{F}^0\), we use the Proxy Anchor (PA) loss \cite{kim2020proxy} to pre-train the model. This PA loss is a deep metric learning loss, which is known for its strong performance and fast convergence. Specifically, it leverages data-to-data and data-to-proxy relationships to minimize the following loss function:
\begin{align}
\ell_{pa}(Z) = &\ \frac{1}{|P^+|} \sum_{p \in P^+} \log \left( 1 + \sum_{z \in Z_p^+} e^{-\alpha (s(z, p) - \delta)} \right) \nonumber \\
&+ \frac{1}{|P^-|} \sum_{p \in P^-} \log \left( 1 + \sum_{z \in Z_p^-} e^{\alpha (s(z, p) + \delta)} \right), \label{eq:pa_loss}
\end{align}
where \(\delta > 0\) is a margin, \(\alpha > 0\) is a scaling factor, \(s(\cdot, \cdot)\) is the similarity function, \(P\) represents the set of all proxies, and \(P^+\) and \(P^-\) are the set of positive and negative proxies in the batch, respectively. For each proxy \(p\), the batch of embedding vectors \(Z\) is split into two sets: \(Z_p^+\) stands for positive embedding vectors of \(p\) and \(Z_p^- = Z \setminus Z_p^+\).

After pretraining, we apply Extreme Value Theory (EVT) to enhance the learned representations and improve the model's ability to detect novel samples. For each proxy \(p_i \in P\), we identify the \(\tau\) nearest samples, \(N_{p_i} = \{(x_j, y_j)\}_{j=1}^{\tau}\), where \(y_{p_i} \neq y_j\), and calculate their distances in the embedding space \(m_{ij} = s(z_j, p_i)\) to the proxy \(p_i\). The distribution of these minimal distances follows a Weibull distribution based on EVT \cite{rudd2017extreme}. The probability of a sample \(x'\) being within the boundary of proxy \(p_i\) is given by the inverse Weibull distribution:
\begin{equation}
\label{eq:psi}
\Psi(p_i, z'; \kappa_i, \lambda_i) = e^{-{\left(\frac{s(p_i, z')}{\lambda_i}\right)}^{\kappa_i}},
\end{equation}
where \(s(p_i, z')\) is the similarity between \(z'\) and proxy \(p_i\), \(z'\) is the embedding vector of sample \(x'\) and \(\kappa_i, \lambda_i\) are the Weibull shape and scale parameters estimated from the smallest \(m_{ij}\) values \cite{rudd2017extreme}. This function provides the probability of sample inclusion (PSI) that is capable of rejecting unknown samples.

So far, EVT has been applied as a post-hoc procedure, meaning it does not affect the learned representation, potentially leading to suboptimal results. To address this, we propose a novel $evt$ loss to fine-tune the model based on the estimated Weibull distribution:
\begin{align}
\label{eq:evt_loss}
\ell_{evt}(Z) &= 
\frac{1}{|P^{+}|} \sum_{p \in P^{+}} \log \left( 1 + \sum_{z \in Z_{p}^{+}} 
\left( 1 - e^{- \left( \frac{s(z, p)}{\lambda_p} \right)^{\kappa_p}} \right) \right) \nonumber \\
&\quad + \frac{1}{|P^{-}|} \sum_{p \in P^{-}} \log \left( 1 + \sum_{z \in Z_{p}^{-}} 
e^{- \left( \frac{s(z, p)}{\lambda_p} \right)^{\kappa_p}} \right), 
\end{align}
where \(\kappa_p, \lambda_p\) are the Weibull shape and scale parameters for proxy \(p\). The first term encourages higher probabilities for positive proxies with respect to positive samples, while the second term penalizes high probabilities for negative proxies. This fine-tuning, in addition to improving the capability of the model for unknown detection, enhances the learned representation compared to the plain PA loss (see Appendix \ref{app:evt_comparison}).

Using Eq. (\ref{eq:psi}), the probability of an input \(x'\) belonging to class \(l\) is computed as: 
\begin{equation}
\label{eq:classifier}
\hat{P}(l \mid z') = \max_{\{i : y_i = l\}} \Psi(p_i, z'; \kappa_i, \lambda_i)
\end{equation}
Using this, the classifier \(\mathcal{C}^0\) is defined as:
\begin{equation}
\label{eq:classifier}
\mathcal{C}^0 =
\begin{cases} 
\arg\max_{l \in \{1, \ldots, M^0\}} \hat{P}(l \mid z'), & \text{if } \hat{P}(l \mid z') \geq \epsilon, \\
\text{``unknown"}, & \text{otherwise,}
\end{cases}
\end{equation}
where \(M^0\) is the total number of classes in the initial stage, and \(\epsilon\) is a threshold for rejecting unknown samples. A practical way to select \(\epsilon\) is to optimize the trade-off between closed-set accuracy and the rejection of unknown classes through cross-class validation. In our experiments, a common threshold worked well enough across different datasets. See Appendix \ref{app:pseudocode:initial} 
for the pseudo-code for training the model in the initial stage.

\subsection{Continual Learning Stage}
After the initial stage and training on the labeled data, the model enters the continual learning stage, where no label will be provided and the input data might contain samples from both novel categories and known categories. The data first goes through a novelty detection module, where it is split into known and unknown samples. The detected unknown samples are then examined using the novel class discovery algorithm to find novel categories. In the end, the model is updated based on these new categories.
\subsubsection{Novelty Detection}
We utilize the classifier, which was trained using EVT in the initial stage to perform novelty detection. More specifically at time \(t\), we use the classifier learned at the previous step \(\mathcal{C}^{t-1}\) to separate unknown samples, as well as assign pseudo labels to known samples
\setlength{\abovedisplayskip}{10pt}
\begin{equation}
\label{eq:classifier_t}
\mathcal{C}^{t-1} =
\begin{cases} 
\arg\max_{l \in \{1, \ldots, M^{t-1}\}} \hat{P}(l \mid z'), & \text{if } \hat{P}(l \mid z') \geq \epsilon, \\
\text{``unknown"}, & \text{otherwise,}
\end{cases}
\end{equation}
\setlength{\abovedisplayskip}{\defaultabovedisplayskip}
where \(M^{t-1}\) is the total number of classes accumulated until time \(t - 1\), and \(\epsilon\) is the threshold for rejecting unknown samples as discussed before. This way the data is split into two sets \(D_\text{known}^{t}\) and \(D_\text{unknown}^{t}\), where \(D_\text{known}^{t}\) is the set of samples belonging to previously known categories (i.e., pseudo labeling) and \(D_\text{unknown}^{t}\) are samples of novel categories (i.e., input for novelty detection), where \(D^t={D_\text{known}^{t} \cup D_\text{unknown}^{t}}\)
\subsubsection{Novel Class Discovery}
The separated unknown samples \(D_\text{unknown}^{t}\) contain potential novel categories that need to be further discovered. To do this the most common approach is clustering these unknown data. Since many clustering algorithms need to know the number of clusters, and estimating this number can add more complexity to the algorithm, we follow \cite{kim2023proxy} by using a non-parametric clustering algorithm. Specifically, we use the affinity propagation method \cite{frey2007clustering} to cluster and discover novel classes. In our experiments, we observed that using affinity propagation for clustering in our proposed scheme leads to over-clustering, which leads to overestimating the number of novel categories, degrading the model's performance in both novel categories and known categories. This problem is mitigated in the following class incremental learning step.
\label{sec:discovery}
\subsubsection{Class Incremental Learning}
The known samples in the data have been assigned pseudo labels by the previous classes and novel samples have their clustering number as pseudo class. Given this, we can integrate the newly discovered categories into the model. To do this we create new proxies for each discovered cluster, i.e., new category and initialize the proxies on the centroids of clusters, following \cite{kim2023proxy}. To achieve this we add the set of proxies of new classes \(P_{new}\) to set of proxies \(P^t = \{P^{t-1} \cup P_{\text{new}}\}\)  and optimize the same proxy anchor loss as in Eq. (\ref{eq:pa_loss}) to improve the performance of the model on the discovered novel categories. 

To avoid catastrophic forgetting during this update, the well-known problem of continual learning, we use feature distillation and feature replay methods. More specifically the distillation loss for feature distillation is defined as 
\begin{align}
\mathcal{L}_{kd}^t(z_o)
&= -\mathbb{E}_{x_o \in D_{\text{known}}^t} \left\| \mathcal{F}^{t-1}(x_o) - \mathcal{F}^t(x_o) \right\|_2, \label{eq:kd_loss}
\end{align}

Additionally, we follow \cite{kim2023proxy} by generating features around each proxy to alleviate catastrophic forgetting. More specifically, for each proxy \(p\), we consider a Gaussian distribution \(\mathcal{N}(p, \sigma^2),\ p \in P^{t-1}\) and generate some features for each category, where the number of generated features is determined based on data balancing. Using the generated features, we define the feature replay loss as follows:
\begin{equation} \label{eq:ex_loss}
\mathcal{L}_{fr}^t(\tilde{Z}) = \mathcal{L}_{pa}^t(\tilde{Z}), \quad 
\tilde{Z} = \{ \tilde{z} \sim \mathcal{N}(p^{t-1}, \sigma^2) \}
\end{equation}

The overall loss for updating the model can be defined as
\begin{equation} \label{eq:total_loss}
\mathcal{L}^t = \mathcal{L}_{pa}^t(Z^t) + \mathcal{L}_{fr}^t(\tilde{Z}) + \mathcal{L}_{kd}^t(z_o)
\end{equation}

where one loss optimizes the model based on the current data for discovered categories and two other losses prevent catastrophic forgetting. 

After training the model on the Eq. (\ref{eq:total_loss}), i.e., getting updated feature extractor \(\mathcal{F}^t\), we derive the classifier \(\mathcal{C}^t\) using the same approach that was used in the initial stage, i.e., using EVT and fitting the Weibull distribution using \(\tau\) nearest points of opposite class samples for each proxy. The overall pseudo code of the continual learning stage is provided in 
Appendix \ref{app:pseudocode:continual}

As was mentioned in section \ref{sec:discovery}, the clustering of novel samples leads to an over-clustering of the data, i.e., a category might have more than one cluster. After training the model on Eq. (\ref{eq:total_loss}), the clusters of the same category are likely to be pushed near each other, we employ this fact to reduce the number of proxies following the approach in \cite{rudd2017extreme} to remove redundant proxies. Let \(p_i\) be a proxy of a discovered novel class and \(\Psi(p_i, p', \kappa_i, \lambda_i)\) be its corresponding fitted Weibull distribution. To decide the redundancy of the pair $\langle p_i, \Psi(p_i, p', \kappa_i, \lambda_i) \rangle$, i.e., whether to keep it, we define an indicator function \(I(\cdot)\) such that
\begin{equation}
\label{eq:indicator}
I(p_i) = 
\begin{cases} 
1 & \text{if } \langle p_i, \Psi(p_i, p', \kappa_i, \lambda_i) \rangle \text{ is kept}, \\
0 & \text{otherwise.}
\end{cases}
\end{equation}
Then, we can define the optimization problem of selecting a minimum number of proxies such that each proxy is at least covered by another proxy as follows:
\begin{equation}
\label{eq:minimize}
\begin{cases}
\text{minimize } \sum_{i=1}^{N} I(p_i) \quad \text{subject to}
\\
\forall j \exists i \mid I(p_i) \Psi(p_i, p_j, \kappa_i, \lambda_i) \geq \zeta.
\end{cases}
\end{equation}
where \(N\) is the number of discovered classes and \(\zeta\) is the threshold used to determine whether a proxy is covered by another proxy. In our experiments, we set this threshold to a very high value near 1 to only reduce proxies that are very close to each other. Since the optimization problem in Eq. (\(\ref{eq:minimize}\)) is a special case of the Karp’s Set Cover problem \cite{rudd2017extreme} and is NP-hard, we follow the greedy approach introduced in \cite{slavik1996tight} to solve this problem approximately. We begin with defining the universe \(U\) as all the proxies, initializing the covered set as an empty set, and finding subsets of covered proxies of each proxy based on Eq.\ref{eq:indicator}. In each iteration, we select the subset that covers the most uncovered proxies, add this subset to the covered set, and repeat this process until all proxies are covered (See 
Appendix \ref{app:pseudocode:reduction}).
\label{sec:continual}

\section{Expermients}
This section provides implementation details and evaluation metrics, and, then, compares the results obtained by the proposed novel method with those obtained through state-of-the-art works. It finally presents the ablation study for the proposed method.
\subsection{Implementation Details}
For the sake of a fair comparison, we utilized ResNet18 \cite{he2016deep}, pre-trained on ImageNet-1k, as the feature extractor across all methods. For the data augmentation, we employed commonly used techniques such as random crops and horizontal flips. For the proxy anchor loss hyperparameters, \(\alpha\) is set to 0.1, and \(\delta\) is set to 32. The hyperparameters \(\tau\), \(\epsilon\), and \(\zeta\) of CATEGORIZER are set to 500, 0.75, and 0.999, respectively. The hyperparameter analysis of \(\tau\) and \(\epsilon\) is shown in Table \ref{tab:neighbours} and Table \ref{tab:novelty_treshold}, respectively. This analysis shows that the sensitivity with respect to these two hyperparameters is low, and a common value has worked well across different datasets. Since \(\zeta\) is only used during the continual stage and can't be configured in the initial stage, we set it to a high value to make it robust across different datasets.
\begin{table}[h!]
\caption{Initial accuracy of the model using different number of neighbours to be used in the EVT modeling. The results show low sensitivity towards this hyperparameter.}
\vskip 0.15in
\begin{center}
\begin{small}
\begin{sc}
\begin{tabular}{l c c c c}
\toprule
Neighbors & CUB & MIT & Dogs & Cars  \\
\midrule
100 & 80.72 & 72.54 & 84.86 & 67.85 \\
250 & 80.91 & 72.45 & 84.70 & 67.86 \\
500 & 80.98 & 72.83 & 84.77 & 68.30 \\
1000 & 80.96 & 72.26 & 84.67 & 68.21 \\
2000 & 80.76 & 72.54 & 84.76 & 68.13 \\
\bottomrule
\end{tabular}
\end{sc}
\end{small}
\end{center}
\vskip -0.1in
\label{tab:neighbours}
\end{table}

\setlength{\tabcolsep}{6pt} 
\begin{table}[h!]
\caption{Accuracy of the novelty detection module using different thresholds. The result indicates low sensitivity towards this hyperparameter}
\vskip 0.15in
\begin{center}
\begin{small}
\begin{sc}
\begin{tabular}{l c c c c}
\toprule
Threshold & CUB & MIT & Dogs & Cars  \\
\midrule
0.3 & 63.76 & 63.82 & 61.64 & 70.19 \\
0.5 & 69.69 & 66.19 & 65.55 & 72.24 \\
0.75 & 70.57 & 67.20 & 68.74 & 75.32 \\
0.95 & 66.03 & 63.67 & 65.05 & 69.83 \\
\bottomrule
\end{tabular}
\end{sc}
\end{small}
\end{center}
\vskip -0.1in
\label{tab:novelty_treshold}
\end{table}

\begin{figure*}
    \centering
    \begin{subfigure}{0.24\linewidth}
        \centering
        \includegraphics[width=1\textwidth]{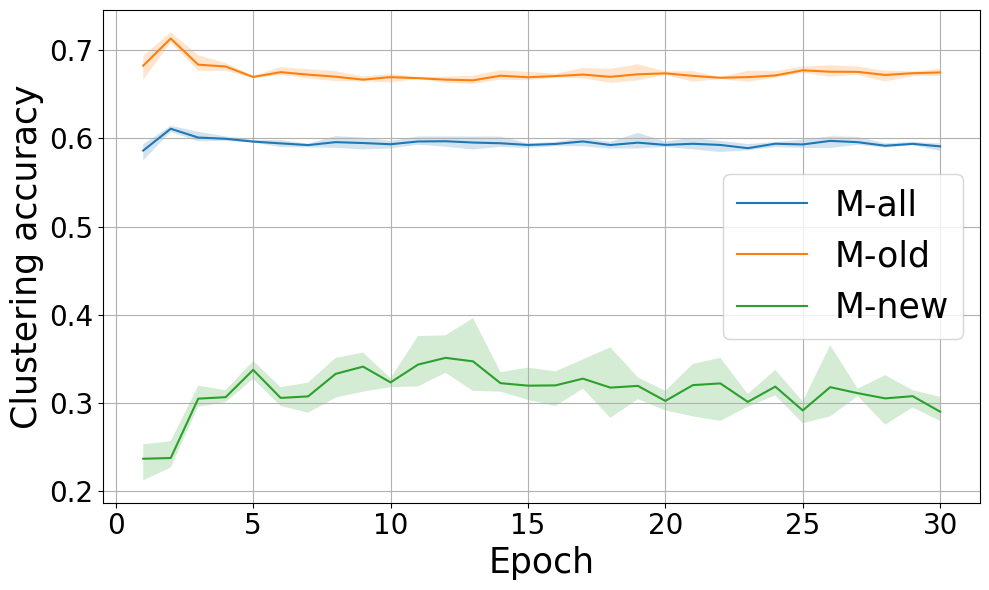}
        \caption{CUB-200}
        \label{fig:cub_accuracy}
    \end{subfigure}
    \begin{subfigure}{0.24\linewidth}
        \centering
        \includegraphics[width=1\textwidth]{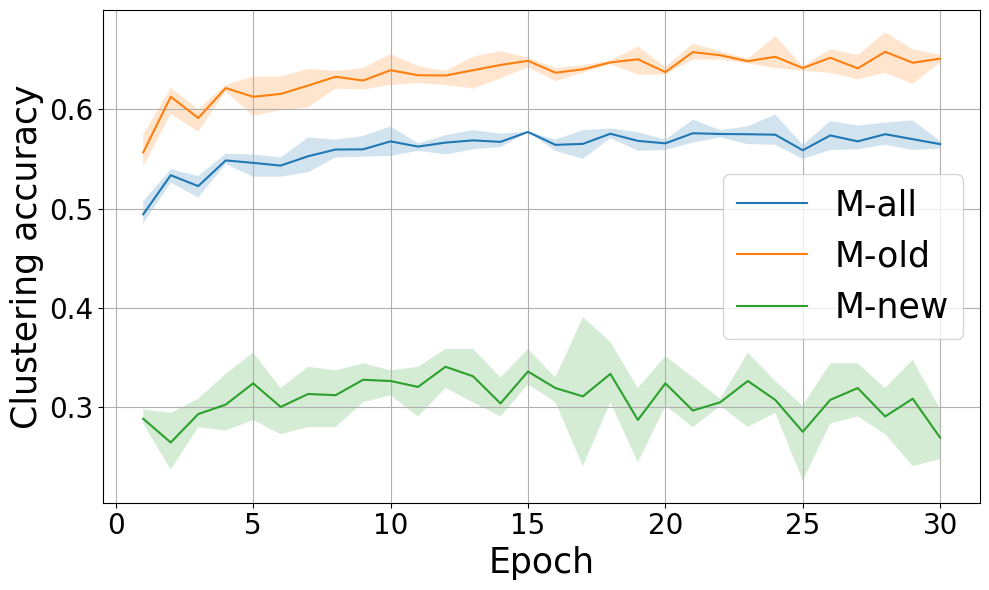}
        \caption{MIT67}
        \label{fig:mit_accuracy}
    \end{subfigure}
    \begin{subfigure}{0.24\linewidth}
        \centering
        \includegraphics[width=1\textwidth]{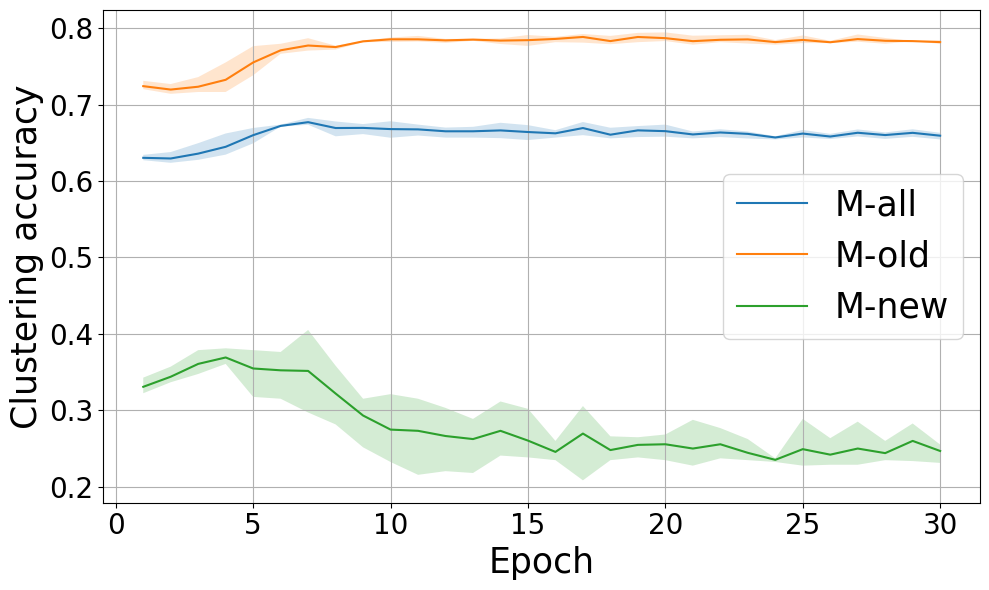}
        \caption{Stanford Dogs}
        \label{fig:dogs_accuracy}
    \end{subfigure}
    \begin{subfigure}{0.24\linewidth}
        \centering
        \includegraphics[width=1\textwidth]{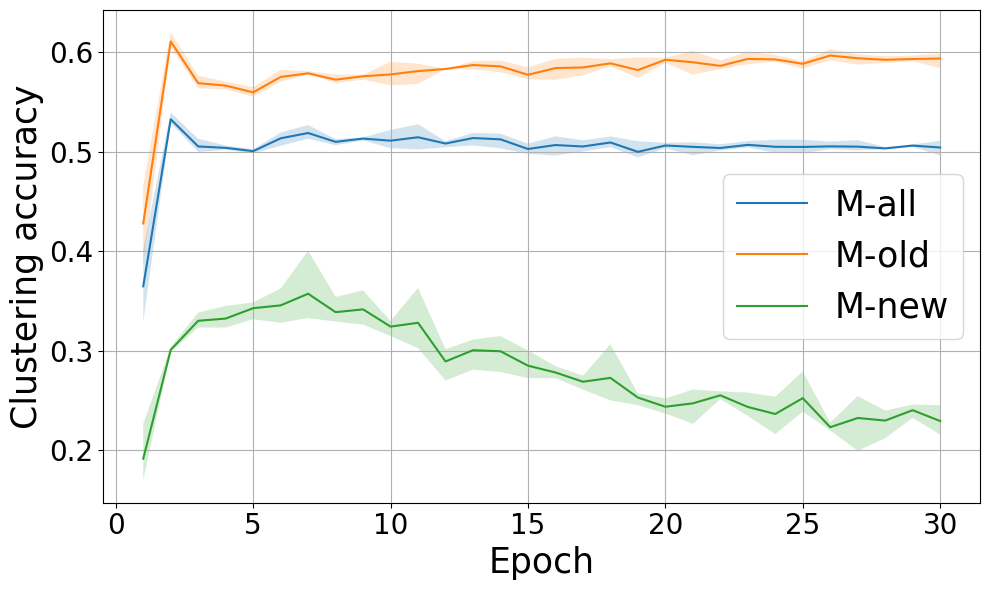}
        \caption{Stanford Cars}
        \label{fig:cars_accuracy}
    \end{subfigure}
    \caption{Clustering accuracy versus epoch number in the continual learning stage over all runs. In the Dogs and Cars datasets, training for longer epochs has led to feature collapse of newly discovered classes, while in CUB and MIT datasets, the accuracy has not changed after certain epochs. Based on this observation we limit the number of epochs of the model training in the continual learning stage.}
    \label{fig:accuracy}
\end{figure*}

We used an  RTX 3070 GPU, Ryzon 5 7600x CPU, and 32GB RAM for running the experiments.
During the initial training stage, the model was trained using the AdamW optimizer with a weight decay of 0.0001 and a learning rate of 0.0001 for 60 epochs on PA loss \cite{kim2020proxy} and 60 epochs on our proposed $evt$ loss. The learning rate was reduced by half every five epochs. In the continual learning stage, the model was updated over 10 epochs. In our experiments, we observed that training beyond this point significantly degrade the performance on novel categories in some datasets (See Figure \ref{fig:accuracy}). To fit the Weibull distributions, we used a torch implementation based on \cite{vast} to estimate the shape and scale parameters of a Weibull distribution.
For other methods, we adhered to the hyperparameters and network architectures specified in their original implementations, referring to the respective papers for details. All reported results represent the average performance over all runs.
\subsection{Evaluation Metrics}
We employ a clustering accuracy-based measurement, called Hungarian assignment algorithm \cite{kuhn1955hungarian}, to measure the performance of the model, following previous methods \cite{zhang2022grow, kim2023proxy,zhao2023incremental, wu2023metagcd}. This metric is defined as
\begin{equation}
\mathcal{M}^t = \frac{1}{|\mathcal{D}|} \sum_{i=1}^{|\mathcal{D}|} \mathbb{I}(y_i = h^*(y_i^*)),
\label{eq:all}
\end{equation}
where \(|\mathcal{D}|\) is the size of the evaluation dataset, $\mathbb{I}$ is the indicator function, $y_i$ and $y_i^*$ are the ground-truth label and clustering assignment for the i-th sample, and \(h^*\) is the optimal assignment. This algorithm aligns predicted clusters with true labels to accurately evaluate the clustering performance. Based on this, we use \(M_{all}\) and \({M_{o}}\) to measure the clustering accuracy on all of the categories and old categories, respectively. To measure the performance drop in the previously known classes after discovering and learning new categories, i.e., forgetting, we employ \(M_f\) metric, which is proposed in \cite{zhang2022grow} and defined as
\begin{equation}
\mathcal{M}_f = \max_t \{\mathcal{M}_o^0 - \mathcal{M}_o^t\},
\label{eq:forgetting}
\end{equation}
where $\mathcal{M}_o^0$ and $\mathcal{M}_o^t$ are the clustering accuracy of old categories, i.e., known categories, at the initial stage and time step \(t\).

To measure the ability of the model to discover and learn novel categories, we employ \(M_d\) measure \cite{zhang2022grow}, which is described as
\begin{equation}
\mathcal{M}_d = \frac{1}{|T|} \sum_{i=T} \mathcal{M}_n^i.
\label{eq:discovery}
\end{equation}
where \(M_n^i\) is the clustering accuracy on novel categories at the \(i\)-th step and \(T\) is the total number of learning steps.
\setlength{\tabcolsep}{10pt} 
\begin{table*}[ht]
\centering
\caption{Comparison of different methods across various datasets w.r.t. various metrics $\mathcal{M}_{all}$, $\mathcal{M}_o$, $\mathcal{M}_f$, and $\mathcal{M}_d$. CATEGORIZER outperforms all other methods in terms of almost all metrics. $\uparrow (\downarrow)$ indicates the metric should have a higher (lower) value.}
\vskip 0.15in
\begin{center}
\begin{small}
\begin{sc}
\begin{tabular}{@{}llcccc@{}}
\toprule
\textbf{Dataset} & \textbf{Method} & $\mathcal{M}_{all} \uparrow$          & $\mathcal{M}_o \uparrow$        & $\mathcal{M}_f \downarrow$ & $\mathcal{M}_d \uparrow$       \\ \midrule
\multirow{6}{*}{CUB} 
& GM   & $12.13 \pm 1.10$ & $13.52 \pm 1.51$ & $37.28 \pm 1.51$ & $10.5 \pm 1.34$ \\
& MetaGCD  & $48.18 \pm 0.68$ & $52.08 \pm 1.10$ & $27.27 \pm 1.10$ & \underline{$32.42 \pm 3.06$} \\
& PA    & $56.52 \pm 0.39$ & $64.32 \pm 0.88$ & $14.39 \pm 0.88$ & $25.77 \pm 3.12$ \\
& IGCD    & \underline{$56.67 \pm 0.43$} & \underline{$66.52 \pm 1.34$} & \textbf{9.55 $\pm$ 1.34} & $17.28 \pm 2.98$ \\ 
& Happy    & \underline{$46.44 \pm 0.43$} & \underline{$54.03 \pm 1.14$} & \textbf{3.27 $\pm$ 1.14} & $16.78 \pm 1.89$ \\ \cmidrule{2-6}
& \textbf{CATEGORIZER} & \textbf{60.82 $\pm$ 0.46} & \textbf{67.72 $\pm$ 0.97} & \underline{12.48 $\pm$ 0.97} & \textbf{33.62 $\pm$ 3.05} \\ \midrule
\multirow{6}{*}{MIT} 
& GM   & $18.50 \pm 1.92$ & $19.37 \pm 1.84$ & $41.68 \pm 1.84$ & $16.93 \pm 1.97$ \\
& MetaGCD  & $40.45 \pm 1.93$ & $43.40 \pm 2.04$ & $31.18 \pm 2.04$ & \underline{$29.29 \pm 2.02$} \\
& PA    & \underline{$52.84 \pm 1.56$} & \underline{$60.19 \pm 1.92$} & \underline{$12.45 \pm 1.92$} & $25.00 \pm 1.97$ \\
& IGCD    & $44.01 \pm 1.75$ & $50.27 \pm 1.99$ & $20.25 \pm 1.99$ & $20.29 \pm 2.32$ \\
& Happy    & \underline{$42.61 \pm 0.24$} & \underline{$51.42 \pm 1.34$} & \textbf{4.90 $\pm$ 1.34} & $9.29 \pm 1.14$ \\ \cmidrule{2-6}
& \textbf{CATEGORIZER} & \textbf{56.42 $\pm$ 1.83} & \textbf{63.11 $\pm$ 2.02} & \textbf{11.70 $\pm$ 2.02} & \textbf{32.86 $\pm$ 2.24} \\ \midrule
\multirow{6}{*}{Dogs} 
& GM   & $11.35 \pm 1.52$ & $13.42 \pm 1.81$ & $44.61 \pm 1.81$ & $10.23 \pm 2.73$ \\
& MetaGCD  & $54.35 \pm 1.23$ & $54.9 \pm 1.45$ & $29.48 \pm 1.45$ & $34.45 \pm 4.42$ \\
& PA    & \underline{$66.23 \pm 1.15$} & \underline{$74.23 \pm 1.95$} & \underline{$10.69 \pm 1.95$} & \underline{$34.69 \pm 4.62$} \\
& IGCD    & $33.63 \pm 0.93$ & $39.15 \pm 1.38$ & $40.11 \pm 1.38$ & $11.52 \pm 2.34$ \\
& Happy    & \underline{$64.75 \pm 0.84$} & \underline{$72.06 \pm 1.22$} & \textbf{5.66 $\pm$ 1.22} & $35.06 \pm 1.99$ \\ \cmidrule{2-6}
& \textbf{CATEGORIZER} & \textbf{68.10 $\pm$ 1.05} & \textbf{76.24 $\pm$ 1.79} & \textbf{7.89 $\pm$ 1.79} & \textbf{36.01 $\pm$ 4.50} \\ \midrule
\multirow{6}{*}{Cars} 
& GM   & $23.52 \pm 2.32$ & $28.24 \pm 2.67$ & $43.26 \pm 2.67$ & $17.23 \pm 1.96$ \\
& MetaGCD  & $47.17 \pm 2.23$ & $52.99 \pm 2.34$ & $23.69 \pm 2.34$ & $24.71 \pm 4.53$ \\
& PA    & $40.46 \pm 2.09$ & $44.35 \pm 2.55$ & $24.30 \pm 2.55$ & \underline{$25.44 \pm 3.62$} \\
& IGCD    & \underline{$48.60 \pm 1.88$} & \underline{$57.82 \pm 2.57$} & \underline{$19.33 \pm 2.57$} & $12.63 \pm 2.75$ \\ 
& Happy    & \underline{$15.06 \pm 0.38$} & \underline{$16.54 \pm 1.27$} & \textbf{9.17 $\pm$ 1.27} & $9.38 \pm 1.28$ \\ \cmidrule{2-6}
& \textbf{CATEGORIZER} & \textbf{51.52 $\pm$ 2.13} & \textbf{58.5 $\pm$ 2.46} & \textbf{18.59 $\pm$ 2.46} & \textbf{32.15 $\pm$ 4.68} \\ \bottomrule
\end{tabular}
\end{sc}
\end{small}
\end{center}
\vskip -0.1in
\label{tab:comparison}
\end{table*}

\setlength{\tabcolsep}{5.6pt} 
\begin{table}[ht]
\centering
\caption{Ablation study on the effect of $evt$ loss and model reduction (RED).}
\vskip 0.15in
\begin{center}
\begin{small}
\begin{sc}
\begin{tabular}{@{}llccccc@{}}
\toprule
\textbf{Dataset} & \(\mathcal{L}_{evt}\) & \textbf{Red} & $\mathcal{M}_{all} \uparrow$          & $\mathcal{M}_o \uparrow$        & $\mathcal{M}_f \downarrow$ & $\mathcal{M}_d \uparrow$       \\ \midrule
\multirow{4}{*}{CUB} 
& \(\times\) & \(\times\)  & $57.03 $ & $65.65$ & $12.70$ & $23.00$ \\
& \checkmark  & \(\times\) & $59.44$ & $66.05$ & $14.15$ & $33.36$ \\
& \(\times\) & \checkmark & $59.39$ & $67.78$ & $10.82$ & $26.28$ \\
& \checkmark & \checkmark & $60.82$ & $67.72$ & $12.48$ & $33.62$ \\
\midrule
\multirow{4}{*}{MIT} 
& \(\times\)   & \(\times\) & $50.93$ & $58.74$ & $13.52$ & $21.36$ \\
& \checkmark & \(\times\)   & $55.82$ & $63.87$ & $10.94$ & $25.71$ \\
& \(\times\) & \checkmark    & $55.72$ & $64.09$ & $10.72$ & $25.00$ \\
& \checkmark & \checkmark    & $56.42$ & $63.11$ & $11.70$ & $31.07$ \\
\midrule
\multirow{4}{*}{Dogs} 
& \(\times\) & \(\times\)  & $62.35$ & $73.12$ & $11.68$ & $23.81$ \\
& \checkmark & \(\times\) & $63.36$ & $72.07$ & $12.06$ & $29.05$ \\
& \(\times\) & \checkmark & $66.40$ & $77.89$ & $6.75$ & $22.09$ \\
& \checkmark & \checkmark & $68.10$ & $76.24$ & $7.89$ & $36.01$ \\
\midrule
\multirow{4}{*}{Cars} 
& \(\times\) & \(\times\)  & $45.92$ & $52.37$ & $16.48$ & $21.05$ \\
& \checkmark & \(\times\) & $47.72$ & $52.80$ & $24.29$ & $29.06$ \\
& \(\times\) & \checkmark & $48.46$ & $55.50$ & $13.32$ & $22.18$ \\
& \checkmark & \checkmark    & $51.52$ & $58.50$ & $18.59$ & $32.15$ \\
\bottomrule
\end{tabular}
\end{sc}
\end{small}
\end{center}
\vskip -0.1in
\label{tab:ablation}
\end{table}
\setlength{\tabcolsep}{3pt} 
\begin{table}[h!]
\caption{Effect of the model reduction on the estimated number of categories. With the model reduction, the estimations are significantly closer to the actual number of categories}
\vskip 0.15in
\begin{center}
\begin{small}
\begin{sc}
\begin{tabular}{l c c c c}
\toprule
Methods & CUB & MIT & Dogs & Cars  \\
\midrule
\# of categories & 200 & 67 & 120 & 196 \\
Estimation w/o reduction & 285 & 134 & 239 & 354 \\
Estimation with reduction & 231 & 81 & 141 & 222 \\
\bottomrule
\end{tabular}
\end{sc}
\end{small}
\end{center}
\vskip -0.1in
\label{tab:reduce}
\end{table}
\subsection{Comparison with State-of-the-art Methods}
We conducted a series of experiments to evaluate CATEGORIZER against state-of-the-art (SOTA) approaches in the CGCD setting. For comparisons, we used the IGCD \cite{zhao2023incremental}, GM \cite{zhang2022grow}, PA \cite{kim2023proxy}, MetaGCD \cite{wu2023metagcd}, and Happy \cite{ma2024happy}, which are very recent works and operate in the same context. Following the experimental setup from \cite{kim2023proxy}, we used 80\% of the classes in the initial stage and introduced the remaining 20\% in the continual learning stage. To better reflect real-world scenarios, where incoming data may include both known and novel classes, only 80\% of the data was utilized during the initial stage, while the remaining 20\% of samples from known classes were mixed with novel samples in the continual learning stage.  

For our experiments, we used fine-grained datasets that closely resemble real-world scenarios: birds \cite{wah2011caltech}, indoor scenes \cite{quattoni2009recognizing}, cars \cite{krause20133d}, and dogs \cite{khosla2011novel}. According to the experimental setup, the initial and continual stage class splits were as follows: birds (160/40), cars (156/40), indoor scenes (53/14), and dogs (96/24). It is important to highlight that our proposed framework is not limited to a specific data type, such as images. It can be applied to any type of data, as long as a suitable backbone network is employed for feature extraction. However, since prior approaches in the CGCD setting have utilized image datasets for their results, we opted to use them as well.

After training the model on the initial stage, we evaluated its accuracy on the evaluation set, as shown in Table \ref{tab:initial_accuracy} in Appendix \ref{app:initial}. CATEGORIZER outperforms others on the CUB200 and MIT67 datasets, while achieving comparable results on the dogs and cars datasets. This table is crucial for analyzing forgetting, as a method with higher initial accuracy might exhibit a higher forgetting rate compared to the one with a lower initial accuracy.  

In addition to accuracy, we evaluated the recall at K metric, to measure the learned representation of different methods in the initial stage as it is essential for novelty detection and overall performance. Results for recall at K (K = 1, 2, 4, and 8) are reported in Appendix \ref{app:evt_comparison}. Our proposed $evt$ loss significantly improves the performance, particularly on the cars dataset, compared to plain PA as shown in Figure \ref{fig:recall_chart} in Appendix \ref{app:evt_comparison}, which reports the Recall@1 performance versus epoch number. Overall, our proposed approach surpasses other methods in terms of representation learning.  

Table \ref{tab:comparison} compares different methods during the continual learning stage. CATEGORIZER consistently outperforms other methods across various datasets and metrics. On the bird's dataset, IGCD exhibited better forgetting performance, likely because its initial accuracy was 4\% lower than ours (See Table \ref{tab:initial_accuracy} in Appendix \ref{app:initial}). GM performs poorly compared to others, which is due to the fact that it requires the ratio of novel category samples on the new dataset. MetaGCD showed good performance in terms of the \(M_d\) metric, demonstrating its ability to learn novel classes effectively, but struggled to retain performance on previously known classes. PA performed well on most datasets, especially dogs, but showed weaker results for the cars dataset, which can be explained by its relatively poor learned representation, as reported in Table \ref{tab:recall} in Appendix \ref{app:evt_comparison}.

\subsection{Ablation Study}
We study the effect of our $evt$ loss and cluster reduction on the performance of the proposed method, which are reported in Table \ref{tab:ablation}.

\textbf{Effectiveness of the $evt$ loss}:
As reported in Table \ref{tab:ablation}, the proposed $evt$ loss enhances performance in terms of most metrics, with a particularly notable improvement in terms of  \(M_d\), highlighting its significance for the novelty detection module. However, on certain datasets, such as Stanford Dogs, the \(M_o\) metric shows reduced performance when using this loss, in exchange for a higher \(M_d\). This indicates a suboptimal trade-off between plasticity and stability, which is mitigated by integrating the $evt$ loss with model-reduction techniques. Additionally, the method exhibits increased forgetting on some datasets due to the $evt$ loss boosting the initial accuracy, making subsequent accuracy drops more pronounced.

\textbf{Effectiveness of model reduction}:
Per Table \ref{tab:ablation}, model reduction introduced in Section \ref{sec:continual} could improve the proposed method in terms of most metrics, particularly the \(M_o\) metric. We believe this improvement is due to the fact that having redundant proxies for the novel categories significantly degrades the performance of old categories. Removing these redundant proxies helps to maintain the model's stability. The combination of the $evt$ loss and model reduction has achieved the most optimal trade-off between maintaining accuracy on the old categories (\(M_o\)) and effective discovery and learning novel categories (\(M_d\)). Table \ref{tab:reduce} shows the effect of the model reduction module on the estimated number of categories in the discovery phase. Upon performing model reduction, the estimated number of categories is significantly closer to the real number of categories.
\setlength{\tabcolsep}{6pt} 
\begin{table}[h!]
\caption{Comparison of novelty detection module of CATEGORIZER and PA. The results show improvement across all the datasets.}
\vskip 0.15in
\begin{center}
\begin{small}
\begin{sc}
\begin{tabular}{l c c c c}
\toprule
Methods & CUB & MIT & Dogs & Cars  \\
\midrule
PA & 59.77 & 60.30 & 65.26 & 71.66 \\
CATEGORIZER & 70.57 & 67.20 & 68.74 & 75.32 \\
\bottomrule
\end{tabular}
\end{sc}
\end{small}
\end{center}
\vskip -0.1in
\label{tab:improved_novelty}
\end{table}

\textbf{Effectiveness on the novelty detection module}: Table \ref{tab:improved_novelty} shows the improvement in the accuracy of the novelty detection module of CATEGORIZER compared to the original PA method \cite{kim2023proxy}. CATEGORIZER has consistently improved the novelty detection accuracy method across different datasets.
\section{Conclusion}
In this paper, we proposed a novel method, called CATEGORIZER, to handle the continual generalized category discovery problem. CATEGORIZER combines proxy anchors and extreme value theory to define decision boundaries around each proxy. We proposed a novel loss, called the $evt$ loss which, enhances the learned representation compared to the plain proxy anchors and outperforms deep metric learning loss used in similar SOTA methods in the CGCD scenario. Furthermore, we mitigated the problem of over-estimating the number of novel categories in the discovery phase by resorting to a novel method, which is based on EVT. CATEGORIZER outperforms state-of-the-art methods in the CGCD scenario across different datasets. In future work, we plan to integrate the $evt$ loss during the continual learning stage of the framework. In addition, we will investigate other clustering methods to be used in the discovery step.
\section*{Impact Statement}
This paper presents work whose goal is to advance the field of Machine Learning. There are many potential societal consequences of our work, none which we feel must be specifically highlighted here.
\nocite{langley00}

\bibliography{example_paper}
\bibliographystyle{icml2025}

\newpage
\appendix
\section{The Method Pseudocodes}
\label{app:pseudocode}
The pseudocode for each part of CATEGORIZER is provided in this appendix. These parts include the initial stage, the overall continual learning stage, and the model reduction algorithm. 
\subsection{Initial Stage}
\label{app:pseudocode:initial}
The pseudocode for the initial stage is provided in Algorithm \ref{alg:initial_stage}. In this stage, the feature is pre-trained using PA loss, followed by fine-tuning on our proposed $evt$ loss.
\begin{algorithm}[H]
\caption{Initial Stage  Training Session.}
\label{alg:initial_stage}
\begin{algorithmic}
\STATE \textbf{Input:} Training dataset \( D^0 \), learning rate \( \eta \), number of epochs \( E \), batch size \( B \), tail size \( \tau \), and threshold \( \epsilon \).

\STATE \textbf{Output:} Feature extractor \( \mathcal{F}^0 \) and classifier \( \mathcal{C}^0 \).
\STATE \textbf{Initialize:} Proxy set \( P \), feature extractor \( \mathcal{F} \) with random weights.
\STATE \textbf{Step 1: Pre-training on PA loss}
\FOR{epoch \( e = 1 \) to \( E \)}
    \FOR{each mini-batch \( \mathcal{B} \subset D^0 \) of size \( B \)}
        \STATE Compute PA loss \( \ell_{pa}(\mathcal{B}) \) (Eq.\ref{eq:pa_loss}).
        \STATE Update feature extractor \( \mathcal{F} \) as:
        \[
        \mathcal{F} \gets \mathcal{F} - \eta \frac{\partial \ell_{pa}(\mathcal{B})}{\partial \mathcal{F}}.
        \]
    \ENDFOR
\ENDFOR

\STATE \textbf{Step 2: Apply EVT}
\FOR{each proxy \( p \)}
    \STATE Estimate Weibull parameters \( \lambda_p \) and \( \kappa_p \) using the tail size \( \tau \).
    \STATE Construct probability of inclusion function \( \Psi \) (Eq.\ref{eq:psi}).
\ENDFOR
\STATE
Construct classifier \(\mathcal{C}^0\) using calculated inclusion functions and threshold \(\epsilon\) (Eq.\ref{eq:classifier})

\STATE \textbf{Step 3: Fine-tuning on $evt$ loss}
\FOR{epoch \( e = 1 \) to \( E \)}
    \FOR{each mini-batch \( \mathcal{B} \subset D^0 \) of size \( B \)}
        \STATE Compute $evt$ loss \( \ell_{evt}(\mathcal{B}) \) (Eq.\ref{eq:evt_loss}).
        \STATE Update feature extractor \( \mathcal{F} \) as:
        \[
        \mathcal{F} \gets\mathcal{F} - \eta \frac{\partial \ell_{evt}(\mathcal{B})}{\partial \mathcal{F}}.
        \]
    \ENDFOR
\ENDFOR

\STATE \textbf{Return} \( \mathcal{F}^0 \text{ and }\mathcal{C}^0 \).
\end{algorithmic}
\end{algorithm}
\subsection{Continual Learning Stage}
\label{app:pseudocode:continual}
The pseudocode for the continual learning stage is provided in Algorithm \ref{alg:continual_learning}. This stage involves detecting unknown samples, discovering novel classes, and integrating the newly discovered classes into the model.
\begin{algorithm}[H]
\caption{Continual Learning Stage.}
\label{alg:continual_learning}
\begin{algorithmic}
\STATE \textbf{Input:} Unlabeled data \( D_u^t \), classifier \( \mathcal{C}^{t-1} \), feature extractor \( \mathcal{F}^{t-1} \), proxy set \( P^{t-1} \), tail size \( \tau \), learning rate \( \eta \), threshold \( \epsilon \), number of epochs \( E \), and batch size \( B \).

\STATE \textbf{Output:} Updated feature extractor \( \mathcal{F}^t \), updated classifier \( \mathcal{C}^t \), and updated proxy set \( P^t \).
\STATE \textbf{Step 1: Split Unlabeled Data}
\STATE Use classifier \( \mathcal{C}^{t-1} \) (Eq.\ref{eq:classifier_t}) to split \( D_u^t \) into:
\[
D_{\text{known}}^t \quad \text{and} \quad D_{\text{unknown}}^t.
\]

\STATE \textbf{Step 2: Process Known and Unknown Data}
\STATE \textbf{Pseudo-labeling:} For \( x \in D_{\text{known}}^t \), assign pseudo-labels using \( \mathcal{C}^{t-1} \).
\STATE \textbf{Clustering:} Apply clustering on \( D_{\text{unknown}}^t \) to group data points into clusters \( C_i \).
\STATE Initialize the proxy set \( P^t \gets P^{t-1} \)
\FOR{each cluster \( C_i \) in \( D_{\text{unknown}}^t \)}
    \STATE Initialize a new proxy \( p_i \) at centroid of \( C_i \) and add it to the proxy set \( P^t \):
    \[
    P^t \gets \{P^t \cup  p_i \}.
    \]
\ENDFOR

\STATE \textbf{Step 3: Train Feature Extractor}
\FOR{epoch \( e = 1 \) to \( E \)}
    \FOR{each mini-batch \( \mathcal{B} \subset D_u^t \) of size \( B \)}
        \STATE Compute the total loss \( \mathcal{L}^t \) as:
        \[
        \mathcal{L}^t = \mathcal{L}_{pa}^t(Z^t) + \mathcal{L}_{fr}^t(\tilde{Z}) + \mathcal{L}_{kd}^t(z_o)
        \]
        (See Eqs.\ref{eq:pa_loss}, \ref{eq:ex_loss}, and \ref{eq:kd_loss}).
        \STATE Update feature extractor \( \mathcal{F}^t \) as:
        \[
        \mathcal{F}^t \gets \mathcal{F}^{t-1} - \eta \frac{\partial \mathcal{L}^t}{\partial \mathcal{F}}.
        \]
    \ENDFOR
\ENDFOR

\STATE \textbf{Step 4: Recalculate Weibull Distributions and Construct Inclusion Functions}
\FOR{each proxy \( p \in P^t \)}
    \STATE Estimate Weibull parameters \( \lambda_p \) and \( \kappa_p \) using tail size \( \tau \).
\ENDFOR


\STATE \textbf{Step 5: Model Reduction}
\STATE Remove redundant proxies of newly discovered categories from the proxy set \( P^t \) based on the updated Weibull distributions (Algorithm~\ref{alg:proxy_reduction}).

\STATE \textbf{Return} Updated feature extractor \( \mathcal{F}^t \), updated classifier \( \mathcal{C}^t \), and updated proxy set \( P^t \).
\end{algorithmic}
\end{algorithm}

\subsection{Model Reduction Stage}
\label{app:pseudocode:reduction}
The pseudocode for the model reduction section of CATEGORIZER, which discards redundant proxies is provided in Algorithm \ref{alg:proxy_reduction}
\begin{algorithm}[h]
\caption{Model Reduction using Greedy Set Cover.}
\label{alg:proxy_reduction}
\begin{algorithmic}
\STATE \textbf{Input:} Proxy set \( P = \{ p_1, p_2, \dots, p_N \} \), corresponding Weibull parameters \( (\lambda_i, \kappa_i) \), inclusion threshold \( \zeta \).
\STATE \textbf{Output:} Reduced proxy set \( P_{\text{reduced}} \).
\STATE \textbf{Step 1: Compute Subsets}
\STATE For each proxy \( p_i \in P \), compute the set of covered proxies (Eq.\ref{eq:minimize}):
\[
S_i = \left\{ p_j \mid I(p_i) \Psi(p_i, p_j, \kappa_i, \lambda_i) \geq \zeta \right\}
\]

\STATE \textbf{Step 2: Initialize Greedy Set Cover}
\STATE Define the universe as all proxies \( U = P \), and subsets as \( \{ S_1, S_2, \dots, S_N \} \).
\STATE Initialize:
\STATE \( \text{covered} \gets \emptyset \)
\STATE \( P_{\text{reduced}} \gets \emptyset \)

\STATE \textbf{Step 3: Greedy Selection}
\WHILE{\( \text{covered} \neq U \)}
    \STATE Find the subset \( S_i \) that covers the maximum uncovered proxies:
    \[
    i^* = \arg\max_{i} \left| S_i \setminus \text{covered} \right|.
    \]
    \STATE Add proxy \( p_{i^*} \) to the reduced proxy set:
    \[
    P_{\text{reduced}} \gets \{ P_{\text{reduced}} \cup p_{i^*} \}.
    \]
    \STATE Update the covered set:
    \[
    \text{covered} \gets \{ \text{covered} \cup S_{i^*}\}.
    \]
\ENDWHILE

\STATE \textbf{Return} Reduced proxy set \( P_{\text{reduced}} \).
\end{algorithmic}
\end{algorithm}

\section{Initial Accuracy}
\label{app:initial}
The initial accuracy of the compared methods is provided in the table \ref{tab:initial_accuracy}.
\begin{figure*}[h!]
    \centering
    \begin{subfigure}{0.24\linewidth}
        \centering
        \includegraphics[width=1\textwidth]{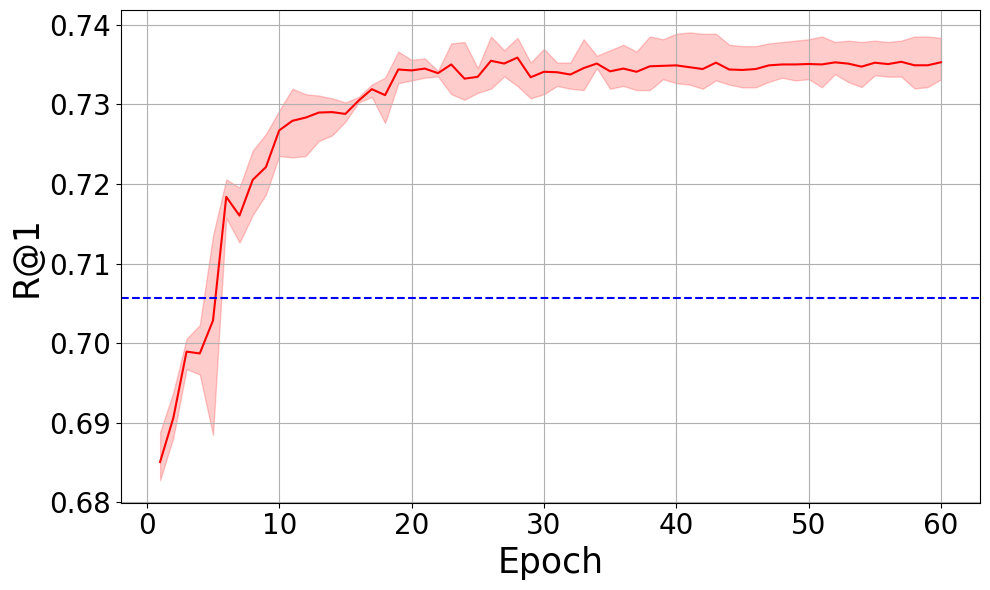}
        \caption{CUB-200}
        \label{fig:cub_recall}
    \end{subfigure}
    \begin{subfigure}{0.24\linewidth}
        \centering
        \includegraphics[width=1\textwidth]{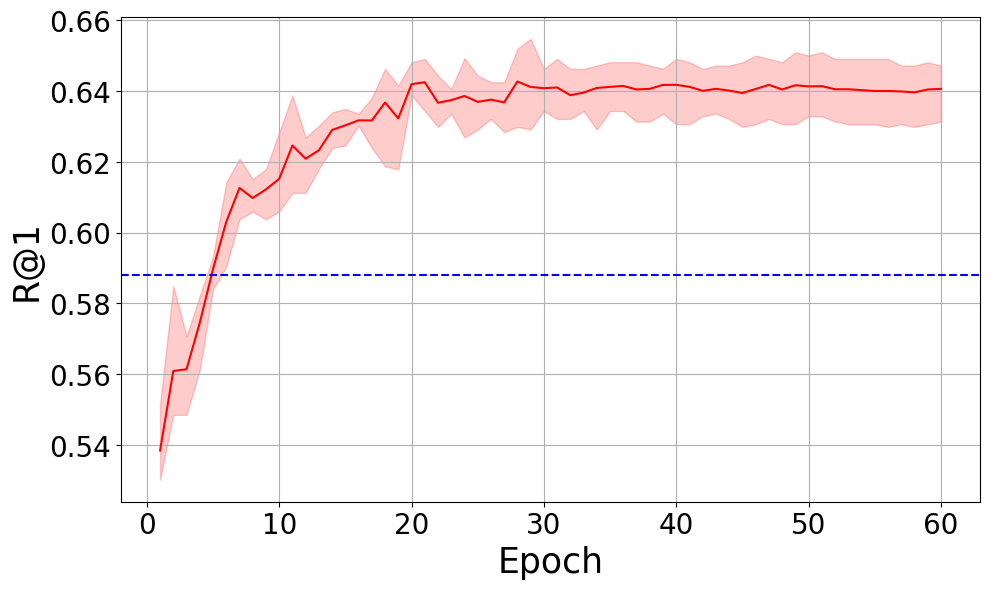}
        \caption{MIT67}
        \label{fig:cub_recall}
    \end{subfigure}
    \begin{subfigure}{0.24\linewidth}
        \centering
        \includegraphics[width=1\textwidth]{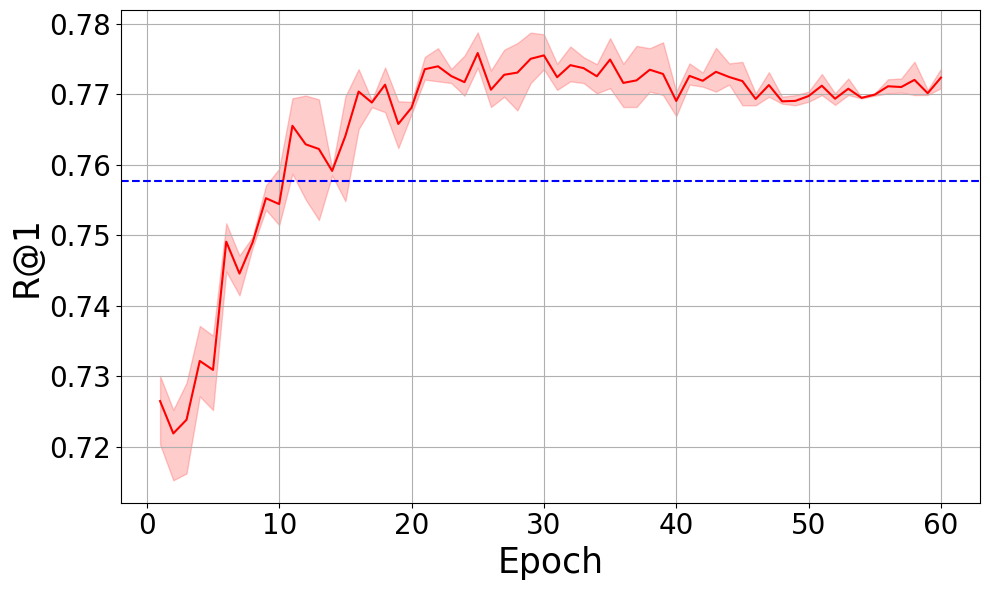}
        \caption{Stanford Dogs}
        \label{fig:cub_recall}
    \end{subfigure}
    \begin{subfigure}{0.24\linewidth}
        \centering
        \includegraphics[width=1\textwidth]{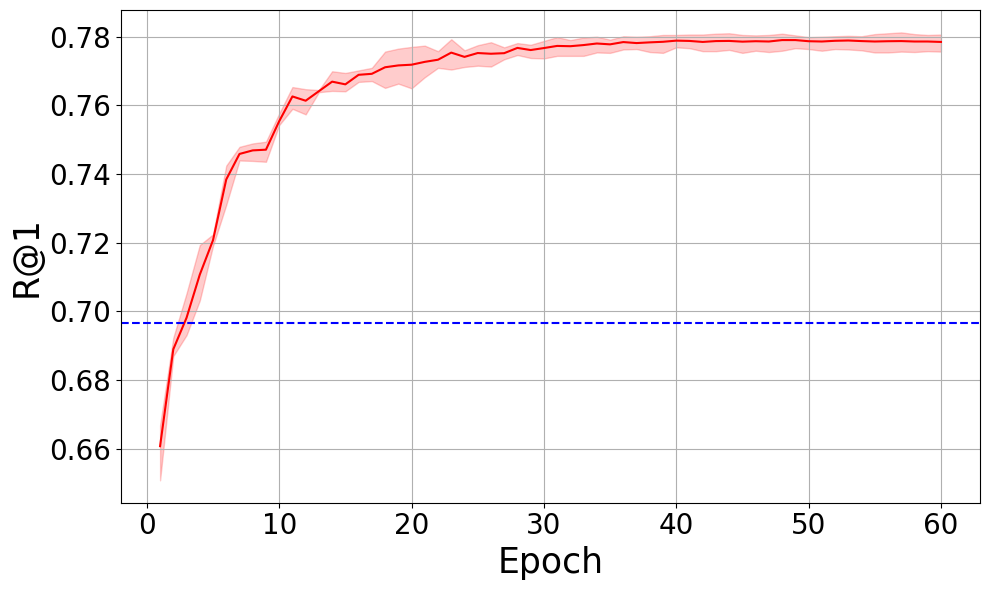}
        \caption{Stanford Cars}
        \label{fig:cub_recall}
    \end{subfigure}
    \caption{Performance in Recall@1 versus epoch number for fine-tuning on our $evt$ loss over all runs. The blue dotted line indicates PA accuracy, i.e., the initial accuracy of the model. The accuracy drops at the beginning of training, bouncing back after a few epochs across different datasets, converging after 10-20 epochs and achieving higher performance compared to that of PA.}
    \label{fig:recall_chart}
\end{figure*}
\begin{table}[H]
\centering
\caption{Comparison of different methods w.r.t. the $\mathcal{M}_o$ metric across various datasets.}
\vskip 0.15in
\begin{center}
\begin{small}
\begin{sc}
\begin{tabular}{lcccc}
\toprule
Methods & CUB & MIT & Dogs & Cars  \\
\midrule
GM & 50.8 & 61.05 & 57.6 & 71.5 \\
MetaGCD & 79.34 & 74.58 & \underline{84.38} & 76.68 \\
PA & 78.71 & 72.64 & \textbf{84.92} & 68.65 \\
IGCD & 76.07 & 70.52 & 79.26 & \textbf{77.15} \\
Happy & 57.30 & 56.32 & 77.72 & \textbf{25.71} \\
Ours & \textbf{80.2} & \textbf{74.81} & 84.13 & \underline{77.09} \\
\bottomrule
\end{tabular}
\end{sc}
\end{small}
\end{center}
\vskip -0.1in
\label{tab:initial_accuracy}
\end{table}

\section{Performance Comparison of the $evt$ Loss}
\label{app:evt_comparison}
The performance comparison of our proposed $evt$ loss compared to other deep metric learning approaches used in similar methods is provided in Table \ref{tab:recall}. Our proposed $evt$ loss outperforms other methods in terms of representation learning. The performance in recall@1 versus epoch number is shown in Figure \ref{fig:recall_chart}.

\setlength{\tabcolsep}{3.6pt} 
\begin{table}[ht]
\centering
\caption{Performance comparison w.r.t. the Recall@K metric on the model trained with different methods. GM \cite{zhang2022grow} uses traditional cross-entropy loss, PA \cite{kim2023proxy} uses proxy anchor loss \cite{kim2020proxy}, and IGCD \cite{zhao2023incremental} and MetaGCD \cite{wu2023metagcd} combine a supervised contrastive loss from \cite{khosla2020supervised} and unsupervised contrastive loss from \cite{chen2020simple} and \cite{gutmann2010noise}, respectively. Our proposed evt loss outperforms all mentioned SOTA methods in almost all datasets.}
\vskip 0.15in
\begin{center}
\begin{small}
\begin{sc}
\begin{tabular}{@{}llcccc@{}}
\toprule
\textbf{Dataset} & \textbf{Method} & R@1          & R@2        & R@4 & R@8       \\ \midrule
\multirow{6}{*}{CUB} 
& GM   & $55.83$ & $67.36$ & $75.80$ & $83.20$ \\
& MetaGCD  & $59.61$ & $69.91$ & $78.97$ & $83.20$ \\
& PA    & \underline{$70.57$} & \underline{$79.47$} & \underline{$86.50$} & \underline{$91.16$} \\
& IGCD    & $64.01$ & $73.85$ & 81.74& $87.36$ \\ \cmidrule{2-6}
& \textbf{CATEGORIZER} & \textbf{73.69} & \textbf{81.17} & \underline{87.22} & \textbf{91.66} \\ \midrule
\multirow{6}{*}{MIT} 
& GM   & $54.70$ & $65.37$ & $73.06$ & $82.23$ \\
& MetaGCD  & $59.32$ & $70.82$ & $78.50$ & $84.47$ \\
& PA    & $58.80$ & $69.70$ & \underline{$79.70$} & \underline{$87.01$} \\
& IGCD    & \underline{$63.06$} & \underline{$72.01$} & $79.10$ & $84.77$ \\ \cmidrule{2-6}
& \textbf{CATEGORIZER} & \textbf{65.47} & \textbf{73.77} & \textbf{80.56} & \textbf{88.01} \\ \midrule
\multirow{6}{*}{Dogs} 
& GM   & $48.20$ & $61.71$ & $72.64$ & $80.49$ \\
& MetaGCD  & $62.65$ & $74.15$ & $83.72$ & $89.16$ \\
& PA    & \underline{$75.77$} & \underline{$84.06$} & \textbf{91.42} & \textbf{95.60} \\
& IGCD    & $64.45$ & $74.46$ & $82.02$ & $87.48$ \\ \cmidrule{2-6}
& \textbf{CATEGORIZER} & \textbf{77.55} & \textbf{84.96} & \underline{90.23} & \underline{93.49} \\ \midrule
\multirow{6}{*}{Cars} 
& GM   & $72.60$ & $81.27$ & $87.50$ & $91.64$ \\
& MetaGCD  & $59.44$ & $71.57$ & $81.33$ & $88.39$ \\
& PA    & $69.66$ & $79.12$ & $86.71$ & $92.42$ \\
& IGCD    & \underline{$77.08$} & \underline{$85.13$} & \underline{$90.41$} & \underline{$94.03$} \\ \cmidrule{2-6}
& \textbf{CATEGORIZER} & \textbf{78.02} & \textbf{85.64} & \textbf{91.08} & \textbf{94.49} \\ \bottomrule
\end{tabular}
\end{sc}
\end{small}
\end{center}
\vskip -0.1in
\label{tab:recall}
\end{table}
\end{document}